\documentclass[aps,prl,twocolumn,superscriptaddress,10pt]{revtex4-2}

\usepackage{graphicx}
\usepackage{dcolumn}
\usepackage{bm}

\usepackage{amssymb}
\usepackage{mathtools}
\usepackage{graphicx}
\usepackage[colorlinks=true, allcolors=blue]{hyperref}
\usepackage{braket}
\usepackage{dsfont}
\usepackage{float}
\usepackage{placeins}

\begin{document}

\title{Local Learning Rules for Out-of-Equilibrium Physical Generative Models}

\author{Cyrill Bösch}
\email{cb7454@princeton.edu}
\affiliation{Department of Computer Science, Princeton University, Princeton, NJ 08540, USA }
\author{Geoffrey Roeder}
\affiliation{Department of Computer Science, Princeton University, Princeton, NJ 08540, USA }
\author{Marc Serra-Garcia}
\affiliation{AMOLF, Science Park 104, 1098 XG Amsterdam, The Netherlands}
\author{Ryan P. Adams}
\affiliation{Department of Computer Science, Princeton University, Princeton, NJ 08540, USA }

\date{\today}

\begin{abstract}
We show that the out-of-equilibrium driving protocol of score-based generative models (SGMs) can be learned via local learning rules.  
The gradient with respect to the parameters of the driving protocol is computed directly from force measurements or from observed system dynamics. 
As a demonstration, we implement an SGM in a network of driven, nonlinear, overdamped oscillators coupled to a thermal bath.
We first apply it to the problem of sampling from a mixture of two Gaussians in 2D. 
Finally, we train an oscillator network on the MNIST dataset to generate images of handwritten digits “0” and “1”.
\end{abstract}
\maketitle

\textit{Introduction}—Physical systems follow complex, stochastic evolution laws, and thus can be seen as special-purpose computers. This paradigm of computation—sometimes referred to as physical computing—has seen a renaissance in recent years \cite{jaeger2023toward, yanagimoto2025programmable, finocchio2024roadmap, melanson2025thermodynamic, aifer2024thermodynamic, duffield2025thermodynamic, lipka2024thermodynamic, anderson2023optical, yildirim2024nonlinear, wanjura2024fully, richardson2025nonlinear, lin2025memristive, nakajima2015information, oguz2024optical}.
While such physical computers are typically designed by solving global optimization problems, the brain works by forming connections based on local information \cite{hebb19490,hopfield1982neural, bengio2015towards}. 
Physical learning seeks to mimic this learning process in synthetic materials \cite{ momeni2024training,stern2020supervised, stern2023learning,hexner2020periodic, scellier2017equilibrium, anisetti2024frequency, stern2021supervised, berneman2025equilibrium, lopez2023self, pourcel2025learning,pourcel2025lagrangian, de2025learning,hexner2023training,wang2024training, laydevant2023benefits,guzman2025unsupervised}. 
This has been successfully implemented in resistor networks \cite{dillavou2022demonstration, dillavou2024machine}, Ising Machines \cite{laydevant2024training, niazi2024training}, physical instantiations of Boltzmann machines \cite{ernoult2019using,kaiser2022hardware, singh2024cmos} and elastic networks \cite{arinze2023learning, altman2024experimental}. 
However, such learning rules typically rely on the notion of an equilibrium or steady state of the system dynamics \cite{scellier2017equilibrium, baldi1991contrastive, stern2021supervised,guzman2025unsupervised}. 
In this Letter, we demonstrate learning in diffusion-based generative models that fundamentally rely on being out-of-equilibrium. 

We introduce ``force matching'', a local learning rule derived from score matching (SM)~\cite{hyvarinen2005scorematching}—the algorithm underlying score-based generative models (SGMs) \cite{song2019generative}. 
Second, we discuss how a variant of contrastive divergence can be used instead, if such force measurements are not accessible. 
In contrast to force matching, this approach requires characterization of physical time scales and high time-resolution observations with respect to those time scales. 
We then numerically train a network of oscillators to implement a generative model—learning the time-dependent, out-of-equilibrium driving protocol required for the system to sample from a target distribution.

Generative modeling is the task of learning to draw samples from a distribution $p$. 
Typically, $p$ is not available in closed form; instead, we have access to it only through a dataset of $M$ samples, ${\mathcal{D} = \{\mathbf{x}^{(m)} \in \mathbb{R}^N\}_{m=1}^M}$.
It poses a fundamental problem in machine learning and statistics, with applications ranging from image generation \cite{song2021scorebased, dockhorn2021score, saharia2022photorealistic} to the discovery of novel proteins \cite{lee2023score}. 
Traditional approaches based on equilibrium sampling (ES), such as Boltzmann machines \cite{hinton1986learning, hinton2002training, salakhutdinov2010efficient}, demonstrated that physical systems can be harnessed to sample from complex distributions through stochastic relaxation. 
Such Boltzmann machines have been realized physically \cite{ernoult2019using,kaiser2022hardware, singh2024cmos} and theorized in chemical \cite{poole2017chemical} molecular \cite{trifonova2025trainable} networks.
However, equilibrium methods typically suffer from long and unpredictable convergence times, as tunneling between isolated high-probability regions is exponentially unlikely~\cite{neal1993probabilistic}. 
Moreover, such Markov processes often traverse low-probability regions, where learned models are less reliable due to data scarcity \cite{song2019generative}.
A promising alternative lies in non-equilibrium processes, where a simple, typically Gaussian distribution is evolved into the target distribution within finite time \cite{song2021scorebased, lipman2022flow, ho2020denoising}. 
These approaches can improve sampling efficiency and make better use of training data by avoiding extended exploration of poorly modeled regions. 
Among the most successful methods are score-based generative models (SGMs), which generate samples by solving non-equilibrium stochastic differential equations (SDEs) \cite{song2019generative,song2021scorebased}.

SGMs frame generation as the process of denoising a sample of pure noise.
A ``forward'' process is designed to add noise to the data and then a ``reverse'' process is learned for removing the noise; this yields a generative model.
Adding noise to the samples of $\mathcal{D}$ is done via a linear SDE \cite{song2021scorebased}. Here we pick \cite{dockhorn2021score}:
\begin{equation}\label{eq:forward}
    \text{d}\mathbf{x} = -\mathbf{x}\,\text{d}t + \sqrt{2k_{\text{B}}T}\,\text{d}\mathbf{w}, \quad \mathbf{x} \in \mathbb{R}^N, \, t \in [0,\tau] 
\end{equation}
where $\text{d}\mathbf{w}$ denotes a Wiener process, $k_{\text{B}}$ is the Boltzmann constant and $T$ is the temperature of the thermal bath.
We assume that the associated probability flow $p_t$, with $p_0 = p$, converges sufficiently well to $p_{\tau} \approx \mathcal{N}(0, k_{\text{B}}T \mathds{1})$ within time $\tau$. 
Note that due to the linearity of the forward SDE, convergence is exponentially fast \cite{vempala2019rapid}.

The backward SDE that reverses $p_t$ is given by \cite{anderson1982reverse}
\begin{multline}\label{eq: backward}
    \text{d}\mathbf{x} =  \big(2k_\text{B}T\nabla_{\mathbf{x}}\log p_t(\mathbf{x}) + \mathbf{x}\big)\,\text{d}t + \sqrt{2k_BT}\,\text{d}\mathbf{w}, \\
    t \in [0,\tau], \quad \mathbf{x}(0) \sim \mathcal{N}(0, k_\text{B}T \mathds{1}),
\end{multline}
where $\nabla_{\mathbf{x}}\log p_t(\mathbf{x})$ is referred to as the ``score''. 
Running this SDE from ${t=0}$ to ${t=\tau}$ generates a sample from the target distribution.
Since $p_t$ is not known, the learning problem becomes that of approximating the score $\nabla_{\mathbf{x}}\log p_t(\mathbf{x})$ using the dataset $\mathcal{D}$. 
Approximating the score, rather than the distribution itself, eliminates the need to estimate the normalization constant—which is equivalent to running the system to equilibrium and is typically intractable for complex, high-dimensional distributions.
Typically, artificial neural networks are used to approximate the score \cite{song2021scorebased}. 
Because the score enters the SDE as a driving term, it can be parameterized in terms of physical forces when the coordinates are interpreted as displacements.
Here, we parameterize the score using the gradient of the energy in a network of conservative, nonlinear, overdamped oscillators.

We assume that the associated energy can be described by a low-order polynomial, both for the local energy, $E^l$, and the coupling energy, $E^c$. 
These energies are parameterized by time-dependent parameters, $\theta_l = \theta_l(t)$ and $\theta_c = \theta_c(t)$.  
Concretely, the local energy is given by
\begin{multline}
    E_{\theta_\text{l}(t)}^{\text{l}}(\mathbf{x}) \\
    := \sum_{n=1}^N \Big[\frac{1}{2}\alpha_n(t)x^2_n + \frac{1}{4}\beta_n(t)x^4_n + \frac{1}{6}\gamma_n(t)x^6_n + f^{ext}_n(t) x_n \Big],
\end{multline}
where we ensure that $\gamma_n(t) > 0$ for all $n$ so that the energy is coercive, and where $\{f^{ext}_n\}_{n=1}^N$ are external forces. The coupling energy of a pair of coupled oscillators is
\begin{multline}
E^{\text{pair}}_{\theta_{nm}(t)}(x_n,x_m) := \frac{1}{2}\kappa_{nm}(t) (x_n-x_m)^2 \\+ \frac{1}{4}\lambda_{nm}(t) (x_n-x_m)^4 
+ \chi_{nm}(t) x_n x_m^2 + \omega_{nm}(t)  x_n^2 x_m,
\end{multline}
where $\kappa$ and $\lambda$ are symmetric linear and quartic couplings, and $\chi$ and $\omega$ are asymmetric couplings that arise, e.g., in optomechanical \cite{sankey2010strong}, magnetic interactions \cite{serra2018tunable} or geometric nonlinearities \cite{serra2016mechanical}. The total coupling energy is then given by
\begin{equation}
    E^{\text{c}}_{\theta_c(t)}(\mathbf{x}) := \sum_{n=1}^N \sum_{m>n}^N E^{\text{pair}}_{\theta_{nm}(t)}(x_n,x_m).
\end{equation}
We define the sum of the local and coupling energy as $E_{\theta(t)} := E^{\text{l}} + E^{\text{c}}$ where $\theta = \{\theta_l,\theta_c\}$. The stochastic dynamics considered here are governed by an overdamped Langevin equation:
\begin{equation}\label{eq:overdamped full}
    \text{d}\mathbf{x} = - \nabla_\mathbf{x} E_{\theta(t)}(\mathbf{x})\,\text{d}t + \sqrt{2k_{\text{B}}T}\,\text{d}\mathbf{w},
\end{equation}
where
\begin{equation}\label{eq:enegy-decomposition}
    E_{\theta(t)} = 2\widehat E_{\theta(t)} - \sum_n x_n^2/2.
\end{equation}

We introduced $\widehat E_{\theta(t)}$ as the part of the energy that is trained such that the associated force satisfies $\hat{\mathbf{f}}_{\theta(t)} = -\nabla_\mathbf{x} \widehat E_{\theta(t)}/k_{\text{B}}T \approx \nabla_\mathbf{x}\log p_t$.
The force that is applied during the reverse process, $\mathbf{f}_{\theta(t)}$, is derived from the total energy $E_\theta(t)$ which includes the unit potential from the forward process (see Eqs.~\eqref{eq:overdamped full}-\eqref{eq:enegy-decomposition} to Eq. \eqref{eq: backward}), i.e. $\mathbf{f}_{\theta(t)} = -\nabla_\mathbf{x} E_{\theta(t)} = -2\nabla_\mathbf{x}\widehat E_{\theta(t)}+\mathbf{x}$.
If the score approximation is sufficiently accurate, then evolving Eq.~\eqref{eq:overdamped full} for time $\tau$ will yield a sample from the target distribution, i.e., $\mathbf{x}(\tau) \sim p$. 
Once trained, the driving protocol is fixed and identical for all reverse trajectories; it is open-loop and requires no measurements or feedback from the evolving state. Hence, the computation is carried out by the physical system.

To ensure that $\mathbf{x}(0) \sim \mathcal{N}(0, k_\text{B}T \mathds{1})$, the system must reach equilibrium before the reverse integration can begin.
This means the system must relax under the force $-{\nabla_\mathbf{x} E_{\theta(\tau)}(\mathbf{x}) = -2\nabla_\mathbf{x}\widehat E_{\theta(\tau)}+\mathbf{x} \approx -\mathbf{x}}$.
However, as noted earlier for the forward process, the system relaxes exponentially fast since the target distribution is a simple Gaussian \cite{vempala2019rapid}. 
The inference process thus involves starting from the equilibrium distribution at $\theta(\tau)$, and then evolving the system under the reverse drive protocol $\theta(\tau-t)$.

Experimental realization of such physical generative models will require changing system parameters over time. Possible avenues include active or robotic matter~\cite{ghatak2020observation,brandenbourger2019non,veenstra2024non}, wave systems in time-varying media~\cite{wang2025experimental,xia2021experimental,galiffi2022photonics, kim2024temporal}, or implementations that emulate time evolution along a spatial dimension~\cite{rechtsman2013strain,mukherjee2020observation}. Electrical circuits are another promising platform for realizing time-varying parameters~\cite{stegmaier2024realizing}; moreover, local learning in electrical circuits has recently been demonstrated beyond the quasi-static limit~\cite{stern2022physical}.

\vspace{1\baselineskip}

\textit{Local learning rule for SM}—
We now tackle the learning problem. 
The SM objective reads \cite{hyvarinen2005scorematching}
\begin{multline}\label{eq:SM loss}
     J_{\text{SM}}(\theta(t))
    =  \mathbb{E}_{\mathbf{x}_t \sim p_t(\mathbf{x})}  \Bigl\{ -\text{Tr}\nabla^2_\mathbf{x}\widehat E_{\theta(t)}(\mathbf{x}_t)/(k_{\text{B}}T) 
    \\+ \frac{1}{2} ||\nabla_\mathbf{x}\widehat E_{\theta(t)}(\mathbf{x}_t)||^2/(k_{\text{B}}T)^2 \Bigl\}.
\end{multline}
The force matching update rule is then obtained by computing the gradient of the objective with respect to the $i$-th parameter:
\begin{multline}\label{eq:localSM}
    \frac{\text{d}J_{\text{SM}}}{\text{d}\theta_i(t)} = \mathbb{E}_{\mathbf{x}_t \sim p_t(\mathbf{x})}\Bigl\{- \text{Tr}\nabla^2_\mathbf{x}\frac{\text{d}\widehat E_{\theta(t)}}{\text{d}\theta_i(t)}/(k_{\text{B}}T)
    \\+ \sum_n \Big[\nabla_\mathbf{x}\frac{\text{d}\widehat E_{\theta(t)}}{\text{d}\theta_i(t)}\Big]_n  \Big[\nabla_\mathbf{x}\widehat E_{\theta(t)}\Big]_n/(k_{\text{B}}T)^2   \Bigr\},
\end{multline}
where we used that the parameter and state derivatives can be exchanged by construction of the energy. 

\begin{figure*}[t!]
\includegraphics[width=1\linewidth]{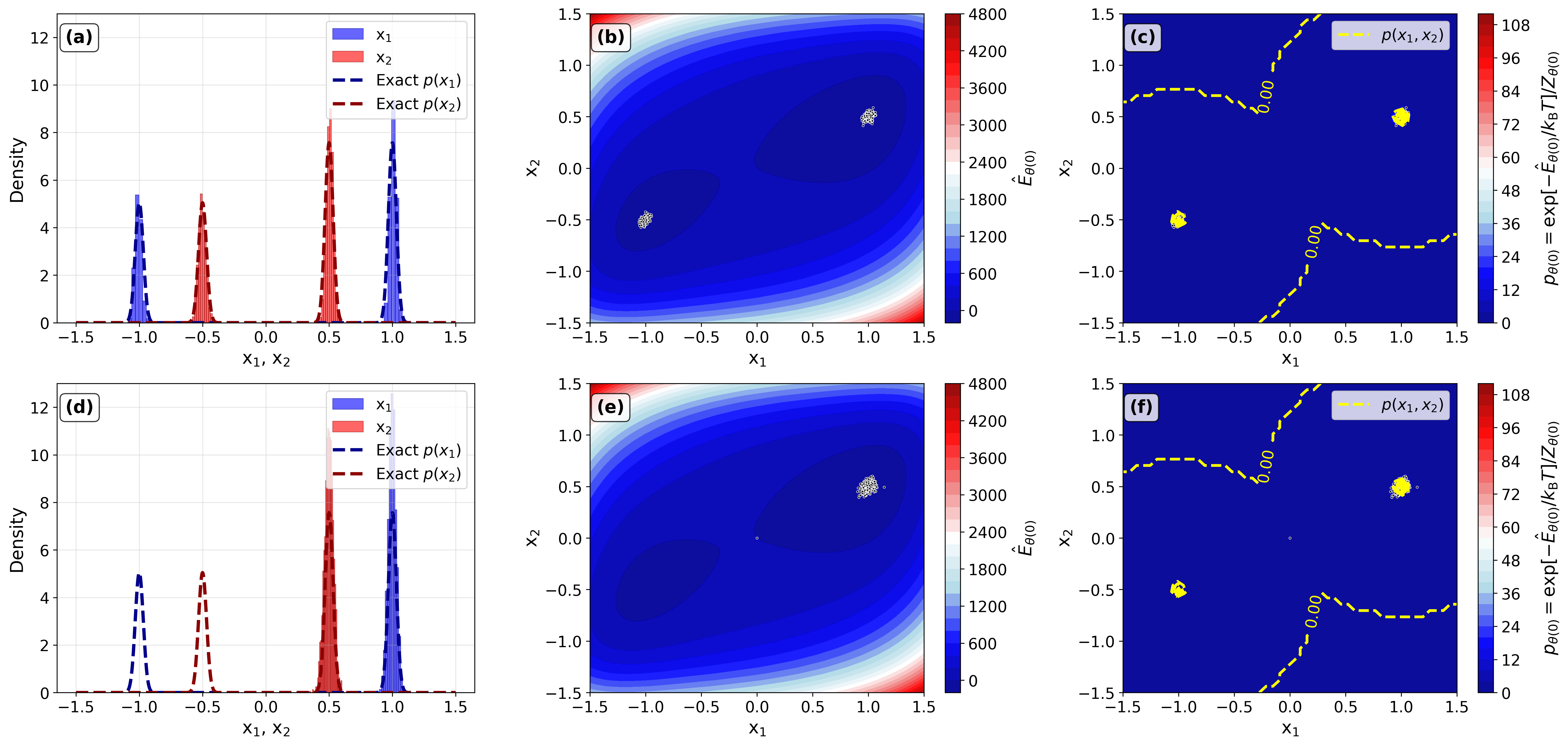}
\caption{(a) $2000$ samples drawn via the learned reverse‐time SDE from a 2-oscillator network overlaid with the exact marginals of $p(x_1,x_2)$.   (b) Learned energy landscape $\widehat E_{\theta(0)}$ at $t=0$ (beginning of the forward, final time of the reverse process), with reverse‐time SDE samples. (c) Iso‐contours of $p(x_1,x_2)$ (red) overlaid on the Boltzmann distribution, $p_{\theta(0)}(x_1,x_2) = \exp[-\widehat E_{\theta(0)}(x_1,x_2)/k_\text{B}T]/Z_{\theta(0)}$, of the learned energy. $Z_{\theta(0)}$ is the partition function. (d)–(f) Corresponding results for ES.}
  \label{fig1}
\end{figure*}

Because any parameter $\theta_i(t)$ enters the energy linearly, $\text{d}\widehat E_{\theta(t)}/\text{d}\theta_i(t)$ is independent of the parameters and can---just as its $x$-derivatives---be evaluated trivially.
Finally, $\nabla_\mathbf{x}\widehat E_{\theta(t)}$ can be identified as the negative force, i.e. $- \hat{\mathbf{f}}_{\theta(t)} := \nabla_\mathbf{x}\widehat E_{\theta(t)} = (\nabla_\mathbf{x} E_{\theta(t)}+\mathbf{x})/2 $. It can be obtained by clamping the network to the sample $\mathbf{x}_t$ and measuring the force acting on each oscillator $\mathbf{f}_{\theta(t)} = -\nabla_\mathbf{x} E_{\theta(t)}$, hence the name ``force matching". Alternatively, if the negative quadratic potential $\sum_n x_n^2/2$ and the factor $2$ in Eq.~\eqref{eq:enegy-decomposition} can be turned off during training, $\hat{\mathbf{f}}_{\theta(t)}$ can be measured directly.

To train the system, we perform the standard approximation of ${p \approx p^{\mathcal{D}}_{t=0} := \sum_{m=1}^M \delta(\mathbf{x} - \mathbf{x}^{(m)}) / M}$ \cite{song2019generative}. 
For the forward process in Eq.~\eqref{eq:forward}, this yields an approximate probability flow $p^{\mathcal{D}}_t \approx p_t$, where $p^{\mathcal{D}}_t$ is a mixture of Gaussians from which we can easily draw samples \cite{song2019generative, dockhorn2021score}. 
We denote the $m$-th sample from $p^{\mathcal{D}}_t$ as $\mathbf{x}^{(m)}_t$.
We discretize time into $N_t$ time points, ${t_0 = 0, \dots, t_{N_t-1} = \tau}$, and solve the minimization of $J_{\text{SM}}$ sequentially. 
To ensure smoothness in the parameter evolution, the learning problem at time $t_{n+1}$ is initialized using the parameters learned at time $t_n$. 
The resulting set of learned parameters, $\{\theta^*(0), \dots, \theta^*(N_t-1)\}$, is then interpolated and applied as the driving protocol at inference time. 
The rate of change in the parameters can be controlled via the temperature: lower temperatures lead to smoother parameter evolution.
As an example, consider the Duffing parameter of the $n$-th oscillator, $\beta_n$, at time $t_l$ and we draw $M$ samples from $p^{\mathcal{D}}_t$:
\begin{multline}
\frac{\text{d} J_\text{SM}}{\text{d}\beta_n(t_l)} = \sum_{m=1}^M \frac{1}{M} \bigl\{ -3(x^{(m)}_{n})^2/(k_{\text{B}}T) \\
- (x^{(m)}_{n})^3 \hat{f}_{n,\theta(t_l)}(\mathbf{x}^{(m)})/(k_{\text{B}}T)^2 \bigr\}.
\end{multline}

In case force measurements are not accessible, the SM objective gradient can instead be obtained from contrastive divergence with one time step (CD1) \cite{hyvarinen2007connections}.
The CD1 loss function is given by
\begin{multline}\label{eq:CD1}
    J_{\text{CD1}}(\theta(t)) =  \mathbb{E}_{\mathbf{x}_t \sim p_t(\mathbf{x}), \mathbf{w}\sim \mathcal{N}(0,\mathds{1})} \\
    \left\{ - \widehat E_{\theta(t)}(\mathbf{x}_t) + \widehat E_{\theta(t)}(\mathbf{x}_{t+\delta}) \right\}.
\end{multline}

The network state evolved by infinitesimal time increment, $\mathbf{x}_{t+\delta}$, can be obtained physically if the quadratic potential and the factor $2$ in Eq.~\eqref{eq:enegy-decomposition} can be turned off during training. Then the network with energy $\widehat E_{\theta(t)}$ can be initialized at $\mathbf{x}_{t}$ and letting it evolve for time $\delta$ will allow us to measure $\mathbf{x}_{t+\delta}$.
In practice the evolution time, $\delta$, must be small compared to the fastest internal time scale.
This must be repeated multiple times to estimate the expectation over noise. 
Formally, we have
\begin{equation}
    \mathbf{x}_{t+\delta} = \mathbf{x}_t - \nabla_\mathbf{x}\widehat E_{\theta(t)}(\mathbf{x}_t)\delta + \sqrt{2\delta k_B T}\mathbf{w},
\end{equation}
where $\mathbf{w} \sim \mathcal{N}(0,\mathds{1})$.

\begin{figure*}[t!]
\includegraphics[width=1\linewidth]{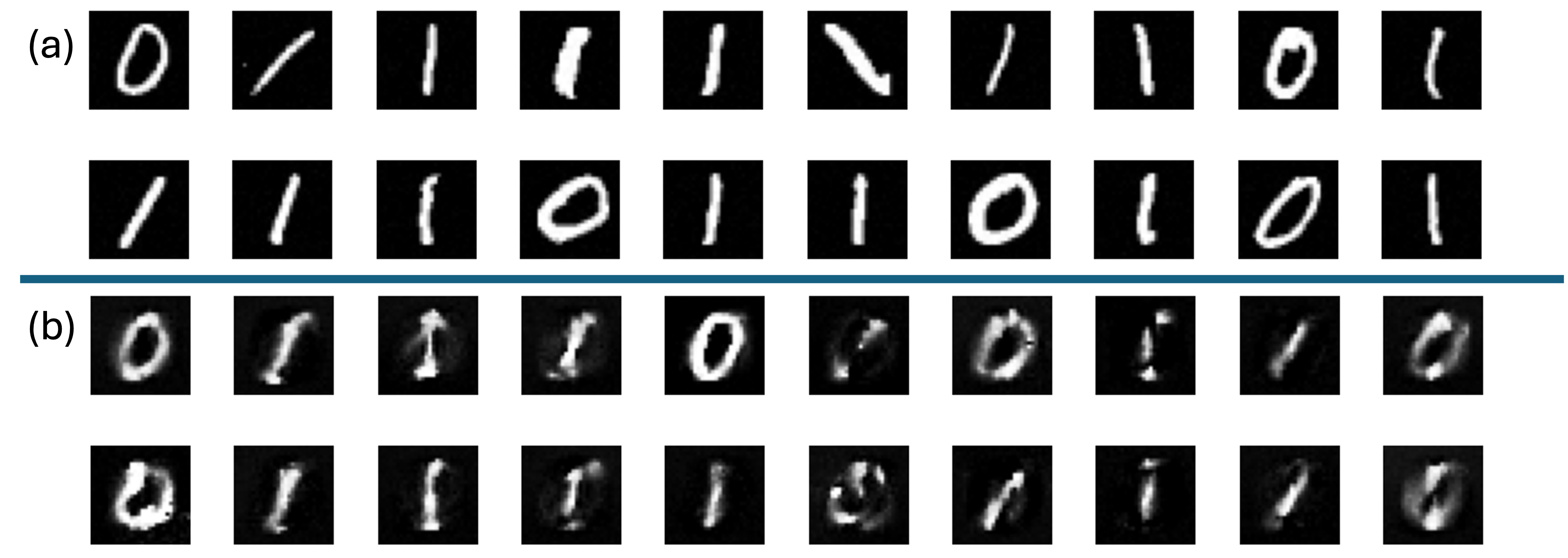}
\caption{(a) Examples of “0”s and “1”s from the MNIST dataset. 
(b) Samples generated with the network nonlinear, overdamped oscillators.}
  \label{fig2}
\end{figure*}

In Hyv\"arinen \cite{hyvarinen2007connections} the correspondence between CD1 and SM was derived for a unit noise strength, which can naturally be extended to finite temperature. 
In the limit where $\delta \rightarrow 0$ we find 
\begin{equation}
   \frac{\text{d}J_{\text{SM}}}{\text{d}\theta_i(t)} = -\frac{1}{\delta (k_\text{B} T)^2}  \frac{\text{d}J_{\text{CD1}}}{\text{d}\theta_i(t)}.
\end{equation}

As in the previous case, since the network parameters enter the energy linearly the derivative $\text{d}J_\text{CD1}/\text{d}\theta$ reduces to a difference of polynomials in the states evaluated at times $t$ and $t+\delta$. 
For the $n$-th Duffing parameter at time $t_l$ we obtain
\begin{equation}
\frac{\text{d} J_\text{CD1}}{\text{d}\beta_n(t_l)} = \sum_{m=1}^M \frac{1}{4M}  \mathbb{E}_{\mathbf{w}\sim \mathcal{N}(0,\mathds{1})} \left\{ -(x^{(m)}_{n,t_l})^4 + (x^{(m)}_{n,t_l+\delta})^4 \right\}.
\end{equation}
Hence, to compute the gradient at each time step $t_l$, the system is initialized multiple times at the same data point $\mathbf{x}^{(m)}_{t_l}$ and evolved for a duration $\delta$ under different noise realizations.
And since the network parameters do not appear in the gradient expression explicitly, no knowledge of the parameter values is required.

\vspace{1\baselineskip}

\textit{Sampling from a mixture of Gaussians and MNIST}—We first apply the presented framework to sample from a mixture of two Gaussians in 2D, i.e., ${p(x_1,x_2) = \omega_1 \mathcal{N}(x_1,x_2|\bm{\mu}_1, \bm{\Sigma}_1) + \omega_2 \mathcal{N}(x_1,x_2|\bm{\mu}_2, \bm{\Sigma}_2)}$. 
Iso-contours are shown in Fig.~\ref{fig1}(c) and (d) in red, and its marginals, $p(x_1)$ and $p(x_2)$, are plotted as dashed red and blue lines, respectively, in Fig.~\ref{fig1}(a) and (d).

To demonstrate the advantages of out-of-equilibrium sampling even in this low-dimensional setting, we compare the samples it produces (first row of Fig.~\ref{fig1}) with those obtained by ES (second row of Fig.~\ref{fig1}). 
We use the force matching learning rule in Eq.~\eqref{eq:localSM} to train a 2-oscillator network. Details on the training are provided in the Supplemental Material. 
Training yields $\widehat E_{\theta(t)}(x_1,x_2)$. 
For the SGM we run Eq.~\eqref{eq:overdamped full} with $-\nabla_\mathbf{x} E_{\theta(t)}(\mathbf{x}) = -2\nabla_\mathbf{x}\widehat E_{\theta(t)}+\mathbf{x}$ for $t \in [0,\tau]$. 
We repeat this procedure $S = 2000$ times to generate$
2000$ samples.

For ES we run the static (i.e., with no time evolution of the parameters) SDE with force
$-\nabla_\mathbf{x}\widehat E_{\theta(0)}$ for a total time $\tau\times S$ and plot the position at every $\tau$-interval.
For sufficiently large $S$, the corresponding Markov chain converges to equilibrium,
i.e., $\mathbf{x}\sim p_{\theta(0)}=\exp[-\widehat E_{\theta(0)}/(k_BT)]/Z_{\theta(0)}\approx p(x_1,x_2)$,
where $Z_{\theta(0)}=\iint \exp[-\widehat E_{\theta(0)}/(k_BT)]\,\text{d}x_1\,\text{d}x_2$.

However, on the simulated time scales the trajectory remains trapped in the high-probability region on the right.
This indicates that ES has not reached equilibrium, despite the fact that the learned energy landscape exhibits two basins.
The reason is that the probability of escaping a basin is exponentially suppressed by the energy barrier separating the two modes.
Moreover, a single transition to the second basin would not be sufficient:
switching must occur sufficiently often such that additional transitions do not change the relative mode weights,
which requires integration times far exceeding those considered here.
In contrast, the SGM generates each sample via a finite-time, out-of-equilibrium process of duration $\tau$
and therefore does not rely on rare barrier-crossing events.

In the Supplemental Material, we also train an oscillator network using the CD1 learning rule to generate samples from a mixture of Gaussians. 
We observe that CD1 gives rise to a slightly different energy.
Although CD1 can still produce high-quality samples in simple settings, force matching proved more reliable overall.
We therefore apply force matching to the larger task of training an oscillator network on the MNIST dataset to generate images of handwritten digits “0” and “1”.
The images are have $28 \times 28$ pixels, corresponding to a network of $28 \times 28$ coupled oscillators. Fig.~\ref{fig2} (a) shows representative training images, and Fig.~\ref{fig2} (b) displays novel samples generated by the trained oscillator network.
The images were generated using long-range couplings between oscillators up to $14$ sites apart. In the Supplemental Material, we study how reducing the number of long-range couplings affects image quality.

\vspace{1\baselineskip}

\textit{Conclusions and outlook}— In this work, we have shown that diffusion models, an out-of-equilibrium system, can be trained with local learning rules. 
In particular, we introduced force matching that allows us to train SGMs on a network of nonlinear, overdamped oscillators coupled to a thermal bath. 
Parameter gradients are obtained directly from force measurements.  
Numerically, we demonstrate that the force matching approach can (i) sample from a mixture of two Gaussians in 2D, and (ii) scale up to a 28 × 28 oscillator network that successfully samples binarized “0”s and “1”s from MNIST.
Alternatively, we showed that CD1 can be harnessed and the updates can be computed from repeated one‐step observations of the physical dynamics.

Future research should study the trade-offs between network size, coupling topology, and sampling performance.  
Furthermore, exploring latent-space SGMs \cite{vahdat2021score} to identify more natural embeddings for the physical system at hand is expected to increase expressivity.
Interestingly, it has recently been shown that locality in the score function can give rise to creativity \cite{kamb2024analytic}, suggesting that what may appear to be a major constraint in physical systems—locality—could in fact be beneficial for generative models.
Hence, our results may inspire novel forms of unconventional computing for generative modeling.

\vspace{1\baselineskip}

\textit{Acknowledgments}— We thank Cindy Zhang for helpful discussions. C.B. was supported by the Swiss National Science Foundation (SNSF) through a Postdoc.Mobility fellowship (P500PT 217673/1).
This work was performed in part at Aspen Center for Physics, which is supported by National Science Foundation grant PHY-2210452.
M.G. was funded by the European Union. Views and opinions expressed are however those of the author(s) only and do not necessarily reflect those of the European Union or the European Research Council Executive Agency. Neither the European Union nor the granting authority can be held responsible for them.
This work is supported by the ERC grant 101040117 (INFOPASS) and NSF OAC-2118201.

\appendix

\section{Training}\label{ap:training}
For the examples shown, we initialized the learning problems as follows: for all $n$, we set $\alpha_n = -1$, $\beta_n = 1$, and $\gamma_n = 1$. All other parameters, including external forces and couplings, were initialized to zero.
In practice, we find that the training dynamics drive $\gamma_n$ to positive values without requiring an explicit constraint. Rarely, in cases of imperfect training (e.g., when a particular coefficient is weakly identified and fluctuates close to zero due to noise/under-training), some $\gamma_n$ can become slightly negative. Since even a single negative $\gamma_n$ makes the corresponding energy contribution non-confining (unbounded from below) and can destabilize the reverse-time integration, we enforce $\gamma_n > 0.001$ during optimization as a technical safeguard to maintain a physically well-defined potential.

We performed a coarse hyperparameter search for each task.
For sampling from a two-dimensional mixture of Gaussians, we found that effective temperatures
$k_\text{B}T = 0.1$ and $k_\text{B}T = 0.01$ yielded the best performance when using the force matching
and contrastive divergence (CD1) learning rules, respectively.
For the MNIST experiments, we found that a temperature of $k_\text{B}T = 0.02$
produced the highest sample quality.

We used $\tau = 5$ for all problems and started the reverse process with samples from $\mathcal{N}(0, k_\text{B}T \mathds{1})$.
We solve the optimization sequentially for $t_0 = 0, \dots, t_{N_t-1} = \tau$, where $N_t = 15$ for the mixture of Gaussians problem and $N_t = 100$ for the MNIST problem.
All training and simulations were performed on a single NVIDIA L40 GPU.

\section{CD1 example}\label{ap:CD1}
\begin{figure*}[ht!]
\includegraphics[width=1\linewidth]{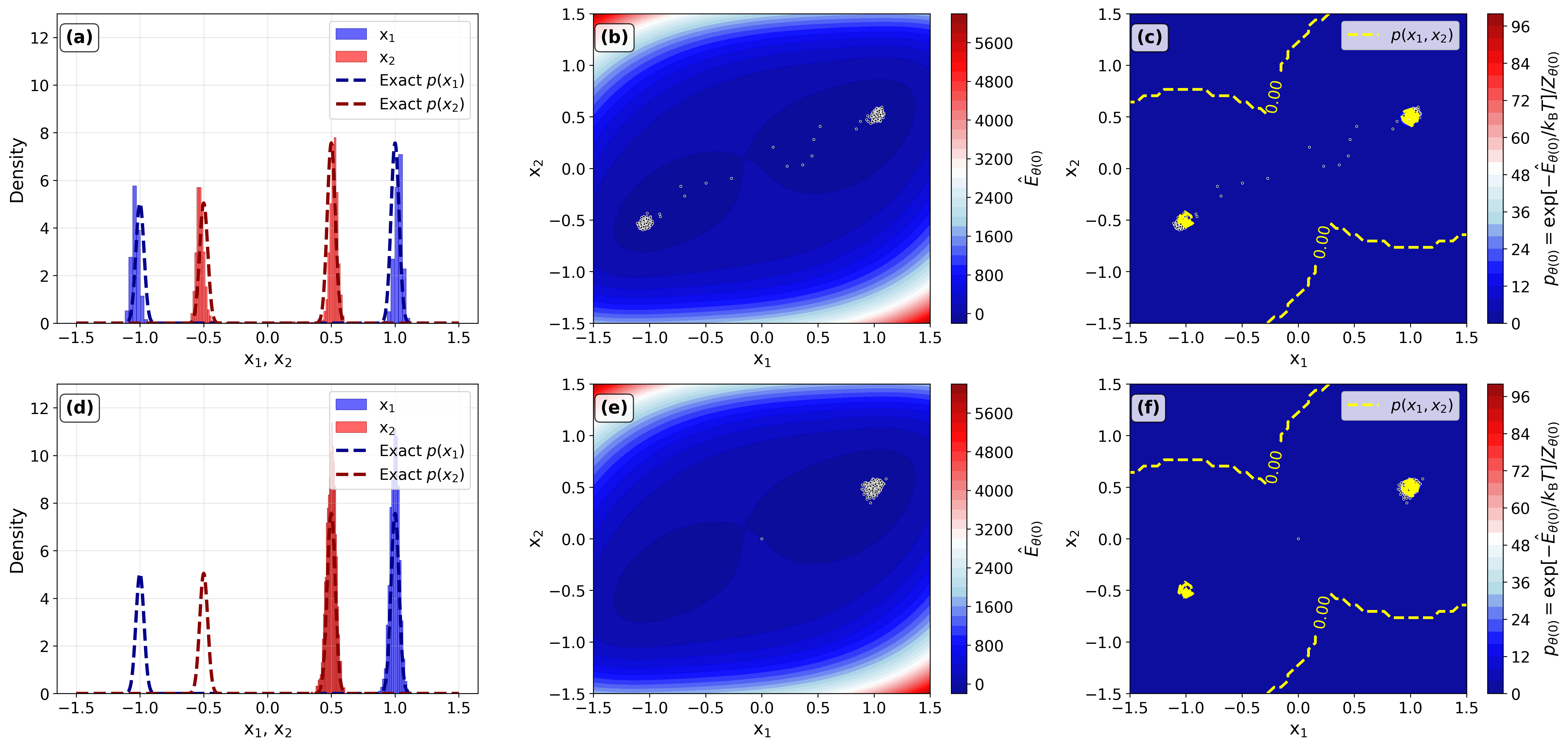}
\caption{Training via CD1 learning rule: (a) $2000$ samples drawn via the learned reverse‐time SDE from a 2-oscillator network overlaid with the exact marginals of $p(x_1,x_2)$.   (b) Learned energy landscape $\widehat E_{\theta(0)}$ at $t=0$ (beginning of the forward, final time of the reverse process), with reverse‐time SDE samples. (c) Iso‐contours of $p(x_1,x_2)$ (red) overlaid on the Boltzmann distribution, $p_{\theta(0)}(x_1,x_2) = \exp[-\widehat E_{\theta(0)}(x_1,x_2)/k_\text{B}T]/Z_{\theta(0)}$, of the learned energy. $Z_{\theta(0)}$ is the partition function. (d)–(f) Corresponding results for ES.}
  \label{fig_CD1}
\end{figure*}

Figure~\ref{fig_CD1} shows an example of a two-oscillator network trained using the CD1 learning rule (see Eq.~(11) in the main text). The resulting energy landscape differs from that obtained with force matching (see Fig.~1 in the main text), yet the generated samples still capture the target distribution. Overall, we find force matching to be more stable.

\section{Number of long-range couplings}\label{a:long-range}
\begin{figure*}
\includegraphics[width=0.85\linewidth]{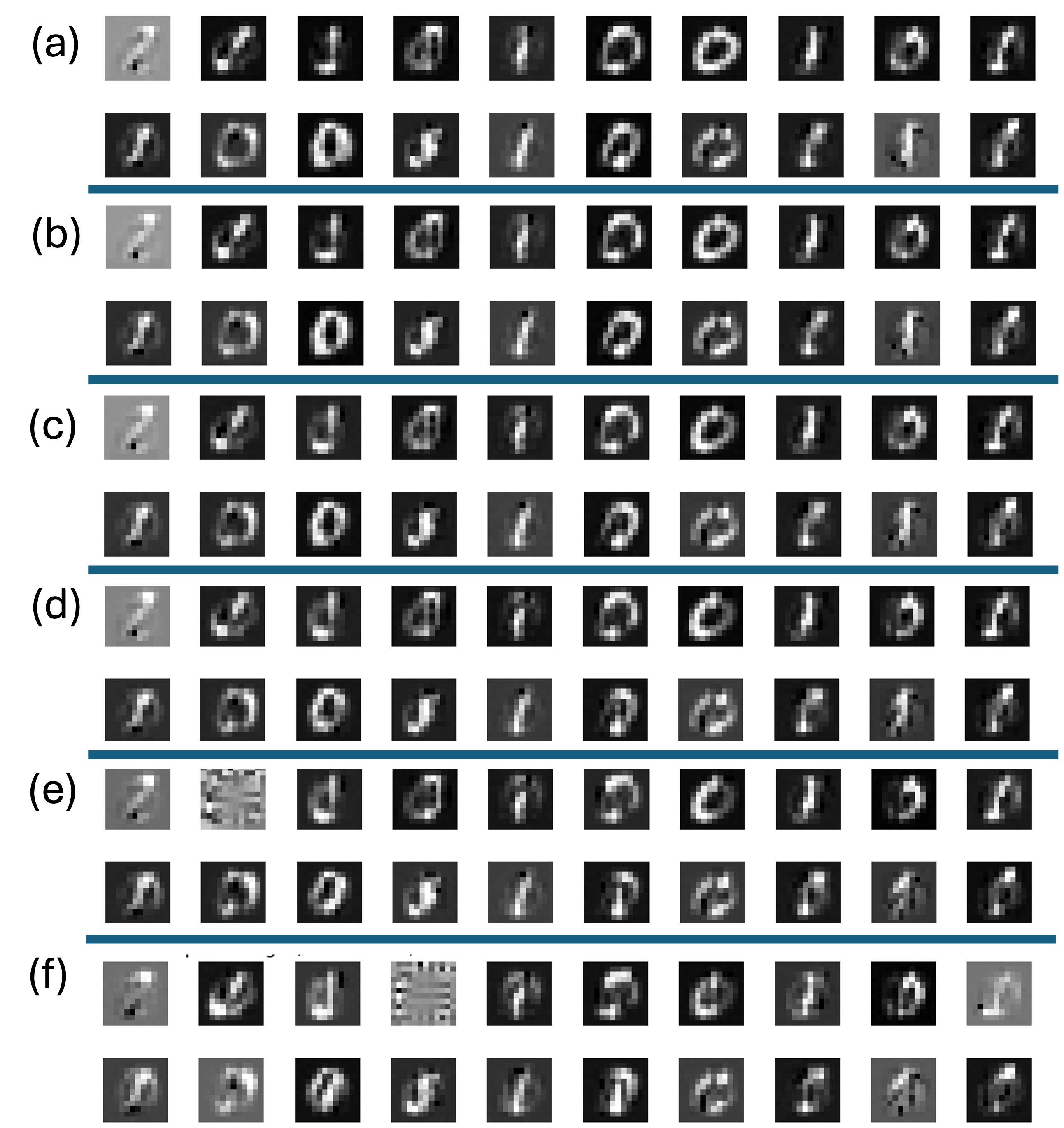}
\caption{Influence of number of couplings on $12 \times 12$ MNIST image generation: $6$ (a), $5$ (b), $4$ (c), $3$ (d), $2$ (e) and $1$ (f) long-range couplings.}
  \label{fig_coupling_influence} 
\end{figure*}

We investigate the effect of long-range couplings on the quality of generated samples (Fig.~\ref{fig_coupling_influence}).
For efficiency reasons we downsampled MNIST to $12\times 12$ pixels.
We progressively reduce the number of long-range couplings from $6$ (Fig.~\ref{fig_coupling_influence}a) to $1$ (Fig.~\ref{fig_coupling_influence}f). 
While the expressive power of the system initially declines slowly, it deteriorates significantly when fewer than $3$ long-range couplings remain.

\FloatBarrier

\bibliography{apssamp}

@article{hyvarinen2005scorematching,
  title={Estimation of Non-Normalized Statistical Models by Score Matching},
  author={Hyv{\"a}rinen, Aapo},
  journal={Journal of Machine Learning Research},
  volume={6},
  pages={695--709},
  year={2005}}

@article{hyvarinen2007connections,
  title={Connections between score matching, contrastive divergence, and pseudolikelihood for continuous-valued variables},
  author={Hyvarinen, Aapo},
  journal={IEEE Transactions on neural networks},
  volume={18},
  number={5},
  pages={1529--1531},
  year={2007},
  publisher={IEEE}
}

@article{song2021scorebased,
  title={Score-Based Generative Modeling through Stochastic Differential Equations},
  author={Song, Yang and Sohl-Dickstein, Jascha and Kingma, Diederik P and Kumar, Abhishek and Ermon, Stefano and Poole, Ben},
  journal={ arXiv:2011.13456},
  year={2021}}

@article{dillavou2022demonstration,
  title={Demonstration of decentralized physics-driven learning},
  author={Dillavou, Sam and Stern, Menachem and Liu, Andrea J and Durian, Douglas J},
  journal={Physical Review Applied},
  volume={18},
  number={1},
  pages={014040},
  year={2022},
  publisher={APS}
}

@article{melanson2025thermodynamic,
  title={Thermodynamic computing system for AI applications},
  author={Melanson, Denis and Abu Khater, Mohammad and Aifer, Maxwell and Donatella, Kaelan and Hunter Gordon, Max and Ahle, Thomas and Crooks, Gavin and Martinez, Antonio J and Sbahi, Faris and Coles, Patrick J},
  journal={Nature Communications},
  volume={16},
  number={1},
  pages={3757},
  year={2025},
  publisher={Nature Publishing Group UK London}
}

@article{aifer2024thermodynamic,
  title={Thermodynamic linear algebra},
  author={Aifer, Maxwell and Donatella, Kaelan and Gordon, Max Hunter and Duffield, Samuel and Ahle, Thomas and Simpson, Daniel and Crooks, Gavin and Coles, Patrick J},
  journal={npj Unconventional Computing},
  volume={1},
  number={1},
  pages={13},
  year={2024},
  publisher={Nature Publishing Group UK London}
}

@article{duffield2025thermodynamic,
  title={Thermodynamic matrix exponentials and thermodynamic parallelism},
  author={Duffield, Samuel and Aifer, Maxwell and Crooks, Gavin and Ahle, Thomas and Coles, Patrick J},
  journal={Physical Review Research},
  volume={7},
  number={1},
  pages={013147},
  year={2025},
  publisher={APS}
}

@article{yanagimoto2025programmable,
  title={Programmable on-chip nonlinear photonics},
  author={Yanagimoto, Ryotatsu and Ash, Benjamin A and Sohoni, Mandar M and Stein, Martin M and Zhao, Yiqi and Presutti, Federico and Jankowski, Marc and Wright, Logan G and Onodera, Tatsuhiro and McMahon, Peter L},
  journal={ arXiv:2503.19861},
  year={2025}
}

@article{finocchio2024roadmap,
  title={Roadmap for unconventional computing with nanotechnology},
  author={Finocchio, Giovanni and Incorvia, Jean Anne C and Friedman, Joseph S and Yang, Qu and Giordano, Anna and Grollier, Julie and Yang, Hyunsoo and Ciubotaru, Florin and Chumak, Andrii V and Naeemi, Azad J and others},
  journal={Nano Futures},
  volume={8},
  number={1},
  pages={012001},
  year={2024},
  publisher={IOP Publishing}
}

@article{wang2024training,
  title={Training coupled phase oscillators as a neuromorphic platform using equilibrium propagation},
  author={Wang, Qingshan and Wanjura, Clara C and Marquardt, Florian},
  journal={Neuromorphic Computing and Engineering},
  volume={4},
  number={3},
  pages={034014},
  year={2024},
  publisher={IOP Publishing}
}

@article{de2025learning,
  title={Learning in a Multifield Coherent Ising Machine},
  author={de Bos, Daan and Serra-Garcia, Marc},
  journal={ arXiv:2502.12020},
  year={2025}
}

@article{scellier2017equilibrium,
  title={Equilibrium propagation: Bridging the gap between energy-based models and backpropagation},
  author={Scellier, Benjamin and Bengio, Yoshua},
  journal={Frontiers in computational neuroscience},
  volume={11},
  pages={24},
  year={2017},
  publisher={Frontiers Media SA}
}

@article{laydevant2024training,
  title={Training an ising machine with equilibrium propagation},
  author={Laydevant, J{\'e}r{\'e}mie and Markovi{\'c}, Danijela and Grollier, Julie},
  journal={Nature Communications},
  volume={15},
  number={1},
  pages={3671},
  year={2024},
  publisher={Nature Publishing Group UK London}
}

@article{hexner2023training,
  title={Training precise stress patterns},
  author={Hexner, Daniel},
  journal={Soft Matter},
  volume={19},
  number={11},
  pages={2120--2126},
  year={2023},
  publisher={Royal Society of Chemistry}
}

@article{oguz2024optical,
  title={Optical diffusion models for image generation},
  author={Oguz, Ilker and Dinc, Niyazi and Yildirim, Mustafa and Ke, Junjie and Yoo, Innfarn and Wang, Qifei and Yang, Feng and Moser, Christophe and Psaltis, Demetri},
  journal={Advances in Neural Information Processing Systems},
  volume={37},
  pages={59150--59173},
  year={2024}
}

@article{lopez2023self,
  title={Self-learning machines based on Hamiltonian echo backpropagation},
  author={Lopez-Pastor, Victor and Marquardt, Florian},
  journal={Physical Review X},
  volume={13},
  number={3},
  pages={031020},
  year={2023},
  publisher={APS}
}

@article{anisetti2024frequency,
  title={Frequency propagation: Multimechanism learning in nonlinear physical networks},
  author={Anisetti, Vidyesh Rao and Kandala, Ananth and Scellier, Benjamin and Schwarz, JM},
  journal={Neural Computation},
  volume={36},
  number={4},
  pages={596--620},
  year={2024},
  publisher={MIT Press One Rogers Street, Cambridge, MA 02142-1209, USA journals-info~…}
}

@article{stern2021supervised,
  title={Supervised learning in physical networks: From machine learning to learning machines},
  author={Stern, Menachem and Hexner, Daniel and Rocks, Jason W and Liu, Andrea J},
  journal={Physical Review X},
  volume={11},
  number={2},
  pages={021045},
  year={2021},
  publisher={APS}
}

@article{stern2023learning,
  title={Learning without neurons in physical systems},
  author={Stern, Menachem and Murugan, Arvind},
  journal={Annual Review of Condensed Matter Physics},
  volume={14},
  number={1},
  pages={417--441},
  year={2023},
  publisher={Annual Reviews}
}

@article{anderson2023optical,
  title={Optical transformers},
  author={Anderson, Maxwell and Ma, Shi-Yuan and Wang, Tianyu and Wright, Logan and McMahon, Peter},
  journal={Transactions on Machine Learning Research},
  year={2023}
}

@article{lee2023score,
  title={Score-based generative modeling for de novo protein design},
  author={Lee, Jin Sub and Kim, Jisun and Kim, Philip M},
  journal={Nature Computational Science},
  volume={3},
  number={5},
  pages={382--392},
  year={2023},
  publisher={Nature Publishing Group US New York}
}

@article{vahdat2021score,
  title={Score-based generative modeling in latent space},
  author={Vahdat, Arash and Kreis, Karsten and Kautz, Jan},
  journal={Advances in neural information processing systems},
  volume={34},
  pages={11287--11302},
  year={2021}
}

@article{dockhorn2021score,
  title={Score-based generative modeling with critically-damped langevin diffusion},
  author={Dockhorn, Tim and Vahdat, Arash and Kreis, Karsten},
  journal={ arXiv:2112.07068},
  year={2021}
}

@article{hinton1986learning,
  title={Learning and relearning in Boltzmann machines},
  author={Hinton, Geoffrey E and Sejnowski, Terrence J and others},
  journal={Parallel distributed processing: Explorations in the microstructure of cognition},
  volume={1},
  number={282-317},
  pages={2},
  year={1986}
}

@article{hinton2002training,
  title={Training products of experts by minimizing contrastive divergence},
  author={Hinton, Geoffrey E},
  journal={Neural computation},
  volume={14},
  number={8},
  pages={1771--1800},
  year={2002},
  publisher={MIT Press}
}

@inproceedings{salakhutdinov2010efficient,
  title={Efficient learning of deep Boltzmann machines},
  author={Salakhutdinov, Ruslan and Larochelle, Hugo},
  booktitle={Proceedings of the thirteenth international conference on artificial intelligence and statistics},
  pages={693--700},
  year={2010},
  organization={JMLR Workshop and Conference Proceedings}
}

@article{song2019generative,
  title={Generative modeling by estimating gradients of the data distribution},
  author={Song, Yang and Ermon, Stefano},
  journal={Advances in neural information processing systems},
  volume={32},
  year={2019}
}

@misc{neal1993probabilistic,
  author       = {Neal, Radford M.},
  title        = {Probabilistic Inference Using Markov Chain Monte Carlo Methods},
  howpublished = {Technical Report CRG-TR-93-1, Department of Computer Science, University of Toronto},
  year         = {1993},
}

@article{lipman2022flow,
  title={Flow matching for generative modeling},
  author={Lipman, Yaron and Chen, Ricky TQ and Ben-Hamu, Heli and Nickel, Maximilian and Le, Matt},
  journal={ arXiv:2210.02747},
  year={2022}
}

@article{ho2020denoising,
  title={Denoising diffusion probabilistic models},
  author={Ho, Jonathan and Jain, Ajay and Abbeel, Pieter},
  journal={Advances in neural information processing systems},
  volume={33},
  pages={6840--6851},
  year={2020}
}

@article{saharia2022photorealistic,
  title={Photorealistic text-to-image diffusion models with deep language understanding},
  author={Saharia, Chitwan and Chan, William and Saxena, Saurabh and Li, Lala and Whang, Jay and Denton, Emily L and Ghasemipour, Kamyar and Gontijo Lopes, Raphael and Karagol Ayan, Burcu and Salimans, Tim and others},
  journal={Advances in neural information processing systems},
  volume={35},
  pages={36479--36494},
  year={2022}
}

@article{jaeger2023toward,
  title={Toward a formal theory for computing machines made out of whatever physics offers},
  author={Jaeger, Herbert and Noheda, Beatriz and Van Der Wiel, Wilfred G},
  journal={Nature communications},
  volume={14},
  number={1},
  pages={4911},
  year={2023},
  publisher={Nature Publishing Group UK London}
}

@article{richardson2025nonlinear,
  title={Nonlinear Computation with Linear Optics via Source-Position Encoding},
  author={Richardson, Nick and B\"osch, C and Adams, Ryan P},
  journal={ arXiv:2504.20401},
  year={2025}
}

@article{wanjura2024fully,
  title={Fully nonlinear neuromorphic computing with linear wave scattering},
  author={Wanjura, Clara C and Marquardt, Florian},
  journal={Nature Physics},
  volume={20},
  number={9},
  pages={1434--1440},
  year={2024},
  publisher={Nature Publishing Group UK London}
}

@article{yildirim2024nonlinear,
  title={Nonlinear processing with linear optics},
  author={Yildirim, Mustafa and Dinc, Niyazi Ulas and Oguz, Ilker and Psaltis, Demetri and Moser, Christophe},
  journal={Nature Photonics},
  volume={18},
  number={10},
  pages={1076--1082},
  year={2024},
  publisher={Nature Publishing Group UK London}
}

@article{hexner2020periodic,
  title={Periodic training of creeping solids},
  author={Hexner, Daniel and Liu, Andrea J and Nagel, Sidney R},
  journal={Proceedings of the National Academy of Sciences},
  volume={117},
  number={50},
  pages={31690--31695},
  year={2020},
  publisher={National Academy of Sciences}
}

@article{vempala2019rapid,
  title={Rapid convergence of the unadjusted langevin algorithm: Isoperimetry suffices},
  author={Vempala, Santosh and Wibisono, Andre},
  journal={Advances in neural information processing systems},
  volume={32},
  year={2019}
}

@article{anderson1982reverse,
  title={Reverse-time diffusion equation models},
  author={Anderson, Brian DO},
  journal={Stochastic Processes and their Applications},
  volume={12},
  number={3},
  pages={313--326},
  year={1982},
  publisher={Elsevier}
}

@article{sankey2010strong,
  title={Strong and tunable nonlinear optomechanical coupling in a low-loss system},
  author={Sankey, Jack C and Yang, Cheng and Zwickl, Benjamin M and Jayich, Andrew M and Harris, Jack GE},
  journal={Nature Physics},
  volume={6},
  number={9},
  pages={707--712},
  year={2010},
  publisher={Nature Publishing Group UK London}
}

@article{serra2016mechanical,
  title={Mechanical autonomous stochastic heat engine},
  author={Serra-Garcia, Marc and Foehr, Andr{\'e} and Moler{\'o}n, Miguel and Lydon, Joseph and Chong, Christopher and Daraio, Chiara},
  journal={Physical review letters},
  volume={117},
  number={1},
  pages={010602},
  year={2016},
  publisher={APS}
}

@article{kamb2024analytic,
  title={An analytic theory of creativity in convolutional diffusion models},
  author={Kamb, Mason and Ganguli, Surya},
  journal={ arXiv:2412.20292},
  year={2024}
}

@article{serra2018tunable,
  title={Tunable, synchronized frequency down-conversion in magnetic lattices with defects},
  author={Serra-Garcia, Marc and Moler{\'o}n, Miguel and Daraio, Chiara},
  journal={Philosophical Transactions of the Royal Society A: Mathematical, Physical and Engineering Sciences},
  volume={376},
  number={2127},
  pages={20170137},
  year={2018},
  publisher={The Royal Society Publishing}
}

@article{ernoult2019using,
  title={Using memristors for robust local learning of hardware restricted Boltzmann machines},
  author={Ernoult, Maxence and Grollier, Julie and Querlioz, Damien},
  journal={Scientific reports},
  volume={9},
  number={1},
  pages={1851},
  year={2019},
  publisher={Nature Publishing Group UK London}
}

@article{kaiser2022hardware,
  title={Hardware-aware in situ learning based on stochastic magnetic tunnel junctions},
  author={Kaiser, Jan and Borders, William A and Camsari, Kerem Y and Fukami, Shunsuke and Ohno, Hideo and Datta, Supriyo},
  journal={Physical Review Applied},
  volume={17},
  number={1},
  pages={014016},
  year={2022},
  publisher={APS}
}

@article{singh2024cmos,
  title={CMOS plus stochastic nanomagnets enabling heterogeneous computers for probabilistic inference and learning},
  author={Singh, Nihal Sanjay and Kobayashi, Keito and Cao, Qixuan and Selcuk, Kemal and Hu, Tianrui and Niazi, Shaila and Aadit, Navid Anjum and Kanai, Shun and Ohno, Hideo and Fukami, Shunsuke and others},
  journal={Nature Communications},
  volume={15},
  number={1},
  pages={2685},
  year={2024},
  publisher={Nature Publishing Group UK London}
}

@article{lin2025memristive,
  title={Memristive linear algebra},
  author={Lin, Jonathan and Barrows, Frank and Caravelli, Francesco},
  journal={Physical Review Research},
  volume={7},
  number={2},
  pages={023241},
  year={2025},
  publisher={APS}
}

@article{nakajima2015information,
  title={Information processing via physical soft body},
  author={Nakajima, Kohei and Hauser, Helmut and Li, Tao and Pfeifer, Rolf},
  journal={Scientific reports},
  volume={5},
  number={1},
  pages={10487},
  year={2015},
  publisher={Nature Publishing Group UK London}
}

@article{bengio2015towards,
  title={Towards biologically plausible deep learning},
  author={Bengio, Yoshua and Lee, Dong-Hyun and Bornschein, Jorg and Mesnard, Thomas and Lin, Zhouhan},
  journal={ arXiv:1502.04156},
  year={2015}
}

@misc{hebb19490,
  title={The organization of behavior},
  author={Hebb, Donald},
  year={1949},
  publisher={New York: Wiley}
}

@article{hopfield1982neural,
  title={Neural networks and physical systems with emergent collective computational abilities.},
  author={Hopfield, John J},
  journal={Proceedings of the national academy of sciences},
  volume={79},
  number={8},
  pages={2554--2558},
  year={1982}
}

@article{baldi1991contrastive,
  title={Contrastive learning and neural oscillations},
  author={Baldi, Pierre and Pineda, Fernando},
  journal={Neural computation},
  volume={3},
  number={4},
  pages={526--545},
  year={1991},
  publisher={MIT Press}
}

@article{dillavou2024machine,
  title={Machine learning without a processor: Emergent learning in a nonlinear analog network},
  author={Dillavou, Sam and Beyer, Benjamin D and Stern, Menachem and Liu, Andrea J and Miskin, Marc Z and Durian, Douglas J},
  journal={Proceedings of the National Academy of Sciences},
  volume={121},
  number={28},
  pages={e2319718121},
  year={2024},
  publisher={National Academy of Sciences}
}

@article{niazi2024training,
  title={Training deep Boltzmann networks with sparse Ising machines},
  author={Niazi, Shaila and Chowdhury, Shuvro and Aadit, Navid Anjum and Mohseni, Masoud and Qin, Yao and Camsari, Kerem Y},
  journal={Nature Electronics},
  volume={7},
  number={7},
  pages={610--619},
  year={2024},
  publisher={Nature Publishing Group UK London}
}

@inproceedings{lipka2024thermodynamic,
  title={Thermodynamic Algorithms for Quadratic Programming},
  author={Lipka-Bartosik, Patryk and Donatella, Kaelan and Aifer, Maxwell and Melanson, Denis and Perarnau-Llobet, Marti and Brunner, Nicolas and Coles, Patrick J},
  booktitle={2024 IEEE International Conference on Rebooting Computing (ICRC)},
  pages={1--13},
  year={2024},
  organization={IEEE}
}

@article{berneman2025equilibrium,
  title={Equilibrium Propagation for Periodic Dynamics},
  author={Berneman, Marc and Hexner, Daniel},
  journal={arXiv preprint arXiv:2506.20402},
  year={2025}
}

@article{momeni2024training,
  title={Training of physical neural networks},
  author={Momeni, Ali and Rahmani, Babak and Scellier, Benjamin and Wright, Logan G and McMahon, Peter L and Wanjura, Clara C and Li, Yuhang and Skalli, Anas and Berloff, Natalia G and Onodera, Tatsuhiro and others},
  journal={arXiv preprint arXiv:2406.03372},
  year={2024}
}

@inproceedings{laydevant2023benefits,
  title={The benefits of self-supervised learning for training physical neural networks},
  author={Laydevant, Jeremie and McMahon, Peter and Venturelli, Davide and Lott, Paul Aaron},
  booktitle={Machine Learning with New Compute Paradigms},
  year={2023}
}

@article{altman2024experimental,
  title={Experimental demonstration of coupled learning in elastic networks},
  author={Altman, Lauren E and Stern, Menachem and Liu, Andrea J and Durian, Douglas J},
  journal={Physical Review Applied},
  volume={22},
  number={2},
  pages={024053},
  year={2024},
  publisher={APS}
}

@article{trifonova2025trainable,
  title={Trainable computation in molecular networks},
  author={Trifonova, Kristina and Falk, Martin J and Rouches, Mason and Vaikuntanathan, Suriyanarayanan and Elowitz, Michael B and Murugan, Arvind},
  journal={bioRxiv},
  pages={2025--12},
  year={2025},
  publisher={Cold Spring Harbor Laboratory}
}

@inproceedings{poole2017chemical,
  title={Chemical boltzmann machines},
  author={Poole, William and Ortiz-Munoz, Andr{\'e}s and Behera, Abhishek and Jones, Nick S and Ouldridge, Thomas E and Winfree, Erik and Gopalkrishnan, Manoj},
  booktitle={International conference on DNA-based computers},
  pages={210--231},
  year={2017},
  organization={Springer}
}

@article{stern2020supervised,
  title={Supervised learning through physical changes in a mechanical system},
  author={Stern, Menachem and Arinze, Chukwunonso and Perez, Leron and Palmer, Stephanie E and Murugan, Arvind},
  journal={Proceedings of the National Academy of Sciences},
  volume={117},
  number={26},
  pages={14843--14850},
  year={2020},
  publisher={National Academy of Sciences}
}

@article{arinze2023learning,
  title={Learning to self-fold at a bifurcation},
  author={Arinze, Chukwunonso and Stern, Menachem and Nagel, Sidney R and Murugan, Arvind},
  journal={Physical Review E},
  volume={107},
  number={2},
  pages={025001},
  year={2023},
  publisher={APS}
}

@article{ghatak2020observation,
  title={Observation of non-Hermitian topology and its bulk--edge correspondence in an active mechanical metamaterial},
  author={Ghatak, Ananya and Brandenbourger, Martin and Van Wezel, Jasper and Coulais, Corentin},
  journal={Proceedings of the National Academy of Sciences},
  volume={117},
  number={47},
  pages={29561--29568},
  year={2020},
  publisher={National Academy of Sciences}
}

@article{brandenbourger2019non,
  title={Non-reciprocal robotic metamaterials},
  author={Brandenbourger, Martin and Locsin, Xander and Lerner, Edan and Coulais, Corentin},
  journal={Nature communications},
  volume={10},
  number={1},
  pages={4608},
  year={2019},
  publisher={Nature Publishing Group UK London}
}

@article{veenstra2024non,
  title={Non-reciprocal topological solitons in active metamaterials},
  author={Veenstra, Jonas and Gamayun, Oleksandr and Guo, Xiaofei and Sarvi, Anahita and Meinersen, Chris Ventura and Coulais, Corentin},
  journal={Nature},
  volume={627},
  number={8004},
  pages={528--533},
  year={2024},
  publisher={Nature Publishing Group UK London}
}

@article{wang2025experimental,
  title={Experimental realization of temporal refraction and reflection in elastic beams},
  author={Wang, Shaoyun and Shao, Nan and Chen, Hui and Chen, Jiaji and Qian, Honghua and Wu, Qian and Duan, Huiling and Al{\'u}, Andrea and Huang, Guoliang},
  journal={Nature Communications},
  volume={16},
  number={1},
  pages={9520},
  year={2025},
  publisher={Nature Publishing Group UK London}
}

@article{stegmaier2024realizing,
  title={Realizing efficient topological temporal pumping in electrical circuits},
  author={Stegmaier, Alexander and Brand, Hauke and Imhof, Stefan and Fritzsche, Alexander and Helbig, Tobias and Hofmann, Tobias and Boettcher, Igor and Greiter, Martin and Lee, Ching Hua and Bahl, Gaurav and others},
  journal={Physical Review Research},
  volume={6},
  number={2},
  pages={023010},
  year={2024},
  publisher={APS}
}

@article{xia2021experimental,
  title={Experimental observation of temporal pumping in electromechanical waveguides},
  author={Xia, Yiwei and Riva, Emanuele and Rosa, Matheus IN and Cazzulani, Gabriele and Erturk, Alper and Braghin, Francesco and Ruzzene, Massimo},
  journal={Physical Review Letters},
  volume={126},
  number={9},
  pages={095501},
  year={2021},
  publisher={APS}
}

@article{galiffi2022photonics,
  title={Photonics of time-varying media},
  author={Galiffi, Emanuele and Tirole, Romain and Yin, Shixiong and Li, Huanan and Vezzoli, Stefano and Huidobro, Paloma A and Silveirinha, M{\'a}rio G and Sapienza, Riccardo and Al{\`u}, Andrea and Pendry, John B},
  journal={Advanced Photonics},
  volume={4},
  number={1},
  pages={014002--014002},
  year={2022},
  publisher={Society of Photo-Optical Instrumentation Engineers}
}

@article{rechtsman2013strain,
  title={Strain-induced pseudomagnetic field and photonic Landau levels in dielectric structures},
  author={Rechtsman, Mikael C and Zeuner, Julia M and T{\"u}nnermann, Andreas and Nolte, Stefan and Segev, Mordechai and Szameit, Alexander},
  journal={Nature Photonics},
  volume={7},
  number={2},
  pages={153--158},
  year={2013},
  publisher={Nature Publishing Group}
}

@article{stern2022physical,
  title={Physical learning beyond the quasistatic limit},
  author={Stern, Menachem and Dillavou, Sam and Miskin, Marc Z and Durian, Douglas J and Liu, Andrea J},
  journal={Physical Review Research},
  volume={4},
  number={2},
  pages={L022037},
  year={2022},
  publisher={APS}
}

@article{guzman2025unsupervised,
  title={Unsupervised and probabilistic learning with Contrastive Local Learning Networks: The Restricted Kirchhoff Machine},
  author={Guzman, Marcelo and Ciarella, Simone and Liu, Andrea J},
  journal={arXiv preprint arXiv:2509.15842},
  year={2025}
}

@article{pourcel2025learning,
  title={Learning long range dependencies through time reversal symmetry breaking},
  author={Pourcel, Guillaume and Ernoult, Maxence},
  journal={arXiv preprint arXiv:2506.05259},
  year={2025}
}

@article{pourcel2025lagrangian,
  title={Lagrangian-based Equilibrium Propagation: generalisation to arbitrary boundary conditions \& equivalence with Hamiltonian Echo Learning},
  author={Pourcel, Guillaume and Basu, Debabrota and Ernoult, Maxence and Gilra, Aditya},
  journal={arXiv preprint arXiv:2506.06248},
  year={2025}
}

@article{mukherjee2020observation,
  title={Observation of Floquet solitons in a topological bandgap},
  author={Mukherjee, Sebabrata and Rechtsman, Mikael C},
  journal={Science},
  volume={368},
  number={6493},
  pages={856--859},
  year={2020},
  publisher={American Association for the Advancement of Science}
}

@article{kim2024temporal,
  title={Temporal refraction in an acoustic phononic lattice},
  author={Kim, Brian L and Chong, Christopher and Daraio, Chiara},
  journal={Physical Review Letters},
  volume={133},
  number={7},
  pages={077201},
  year={2024},
  publisher={APS}
}

\end{document}